\documentclass{article}

\usepackage[utf8]{inputenc} % allow utf-8 input
\usepackage[style=numeric, backend=biber, sorting=none]{biblatex}
\usepackage[preprint]{custom}
\usepackage{amsmath}

\usepackage[T1]{fontenc}    % use 8-bit T1 fonts
\usepackage{hyperref}       % hyperlinks
\usepackage{url}            % simple URL typesetting
\usepackage{booktabs}       % professional-quality tables
\usepackage{amsfonts}       % blackboard math symbols
\usepackage{nicefrac}       % compact symbols for 1/2, etc.
\usepackage{microtype}      % microtypography
\usepackage{xcolor}         % colors
\usepackage[pdftex]{graphicx}
\usepackage[normalem]{ulem}
\usepackage{color}
\usepackage{tabularray}
\usepackage{authblk}

\title{Deep Learning Surrogates for Emulating Stochastic Climate Tipping Dynamics}

\author[1]{Adeline Hillier}
\author[1]{Jennifer Sleeman}
\author[1]{Jay Brett}
\author[1]{Caroline Tang}
\author[1]{Jenelle Millison}
\author[2]{Anand Gnanadesikan}

\affil[1]{Johns Hopkins Applied Physics Laboratory}
\affil[2]{Johns Hopkins University}

\begin{document}

\maketitle

\begin{abstract}
This work explores a dynamics-informed Temporal Fusion Transformer (TFT) as a data-driven surrogate for computationally intensive Earth system simulations. Focusing on multivariate time series describing global ocean transport, we demonstrate the surrogate's ability to forecast tip events across thousands of time steps. The data involve up to 21 non-stationary time series in addition to static covariates describing free parameters and initial conditions. Modifications to the architecture and objective function yield a surrogate that anticipates the timing of Atlantic and Pacific collapses to high fidelity and captures the stochastic uncertainty in transition timing across ensemble predictions. The learned surrogate achieves a $465\times$ computational speedup over the numerical simulator while maintaining differentiability with respect to parameters and initial conditions.
\end{abstract}

\section{Introduction}
In Earth systems and other nonlinear dynamical systems, forecasting sudden state changes is often complicated by sensitivity to initial conditions, potential feedbacks, and the intrinsically chaotic nature of the governing dynamics \parencite{ritchie2021overshooting}.  Small changes in external forcings can result in large non-linear responses that are often irreversible, making reliable prediction of abrupt transitions challenging. In early warning theory, there are typically indicators that a slowing down occurs as a bifurcation is approached \parencite{nazarimehr2020critical}.  These indicators have traditionally been used to forecast tipping points \cite{Bury2021}; however, not all tipping points exhibit this slowing behavior, and some early warning signals may not be defined in terms of bifurcation theory \parencite{lenton2011early}.  For this reason, traditional statistical approaches used to detect early warning signals can be sensitive to a number of attributes that make them limited in terms of their reliability in predicting these abrupt transitions.  

Fully resolved numerical models tend to provide physically consistent representations of climate dynamics and can, in principle, simulate tipping point behavior. However, poorly constrained parameters coupled with significant computational expense limit their utility for ensemble-based uncertainty quantification and parameter inference. Reduced-order models (ROMs) address this computational bottleneck by projecting the system into a lower-dimensional state space that preserves essential dynamical features while reducing the number of prognostic equations \parencite{levermann2010atlantic}. However, even ROMs used to study abrupt shifts remain limited in their ability to systematically characterize tipping behavior \parencite{Sleeman1}. In particular, identifying regions of initial condition space that lead to abrupt transitions is challenging due to the reduced, but still complex search space present. While methods exist to probe such regions \cite{Sleeman2}, exhaustive exploration remains computationally expensive even at the ROM level. As a result, the mechanisms governing tipping behavior---and the conditions under which it occurs---are still not fully understood.

Deep learning provides a complementary, data-driven approach to anticipating regime shifts for both early warning and forecasting applications. There are models that detect proximity to bifurcations by learning early warning signal patterns \parencite{lenton2011early,dakos2024tipping}.  In our study, we target forecasting future trajectories, with a focus on collapses of the Meridional Overturning Circulation (AMOC in the Atlantic; PMOC in the Pacific). We leverage reduced-order models to generate labeled time series spanning multiple dynamical regimes, which we use to train a modified Temporal Fusion Transformer (TFT) as a surrogate emulator. Our objectives are to accurately forecast full trajectories, reproduce the timing and occurrence of AMOC/PMOC collapses, and capture the distribution of collapse times under stochastic forecasting. The TFT architecture is well-suited to this problem because it integrates static covariates (model parameters, initial conditions) with time-varying covariates (state variables, external forcings), learns long-horizon temporal dependencies, and provides interpretable variable selection mechanisms that can inform identification of dynamically relevant covariates.  To address the uncertainty and variability present in the stochastic time series, we modify the learning objective with a time-warping loss \cite{softDTW} that appears to tolerate phase variability while preserving the shape and amplitude of predicted trajectories. We demonstrate that the trained surrogate forecasts collapse timing with high correlation, reproduces the probability distribution of transition times under stochastic forcing, achieves a $465\times$ speedup relative to numerical integration, and maintains differentiability with respect to all input parameters, thereby enabling gradient-based sensitivity analysis and parameter inference.

\section{Background}
The Meridional Overturning Circulation (MOC) involves warm, salty water being transported northward throughout the Atlantic but not the Pacific. In the far Northern Atlantic, the water cools and increases in density, then sinks into the deep ocean before returning southward. The theory links the possibility of abrupt transitions in the overturning circulation in both the Atlantic (AMOC) and the Pacific (PMOC) to the salt-advection feedback, where short-term decreases in salinity in high latitudes weaken the overturning and reduce the supply of warm salty water, amplifying the initial disturbance \parencite{stommel1961thermohaline, jackson2013shutdown}. A collapse of this system would have profound and potentially irreversible consequences on heat and carbon uptake, oxygen distribution, and marine ecosystems. Although much of the research focuses on the North Atlantic in terms of tipping point behavior, global overturning is fundamentally a coupled multi-basin phenomenon \cite{ dijkstra2024effect} involving the Pacific and the Southern Ocean affecting properties such as water-mass transformation and return pathways \parencite{marshall2012closure}.

To analyze overturning as a tipping point, in this work we employ two idealized box models. The six-box model partitions the ocean into connected regions, or “boxes,” representing the high and low latitude Atlantic, high and low latitude Pacific, Southern Ocean and deep ocean \parencite{6BoxModel}. Each box exchanges heat and freshwater with its neighboring boxes, allowing the model to reproduce key processes such as northward transport of warm salty water, sinking of dense waters, and wind-driven upwelling. In parallel, we use a four-box model \parencite{gnanadesikan2018flux} that excludes the Pacific basins, providing a controlled benchmark to study how inter-basin coupling alters stability and sensitivity. 

Although reduced-order box models are simplistic, they still generate complex behaviors, including abrupt shifts, where deep water formation rates and circulation strength change in response to small perturbations in the hydrological cycle. Since the box models enable interpretability while representing the nonlinearity well, they are good candidates for use in this deep learning study.  The models enable the generation of labeled synthetic time series data across circulation regimes, support counterfactual experiments by perturbing external forcings, and provide ground truth both for forecasting shifts and for identifying their causal drivers, which will support future intervention work. We use these reduced-order models to train deep learning surrogates that emulate overturning dynamics and forecast regime shifts while retaining physically meaningful relationships in the underlying system.

\section{Related work}

Producing accurate time-recorded forecasts of critical transitions in noisy chaotic systems is a difficult undertaking. Deep learning models that focus on detecting early warning signals, such as critical slowing down, may correctly flag and classify an impending tip but cannot localize transitions in time \parencite{Bury, Deb}. \textcite{Liu} trained a GIN-GRU architecture to forecast the critical control parameter threshold at which a bifurcation is likely to occur, successfully fine-tuning it on empirical data from unseen districts to forecast rainfall-driven rainforest-savanna transitions with low error. However, their approach assumes a single bifurcation parameter, limiting applicability to real climate tipping points driven by multiple interacting controls. According to \textcite{Bury}, predicting the timing of tipping points would require forecasting system trajectories. Traditional Earth System Models (ESMs) do not provide exact forecasts due to their sensitivity to approximate parameterizations for subgrid-scale physics \parencite{6BoxModel,gnanadesikan2018flux,Irrgang}. Additionally, because abrupt transitions have no analogues in the short instrumental record, the ability of current ESMs to predict such rare, high-risk events remains untested. When it comes to long-horizon bifurcation events, reduced-order models are often more statistically accurate than full-order models \parencite{Behnoudfar}.

Graph neural networks (GNNs) approach long-horizon multivariate time series (MTS) forecasting by making cross-variable structure explicit: a spatial message-passing layer operates over a (known or learned) graph while a temporal module handles sequence dynamics. GNNs elegantly handle a central problem in MTS forecasting: that of capturing dependencies across variables, sometimes referred to as channel or feature fusion. \textcite{Cini} use powers of a graph shift operator to propagate spatial information between embeddings; the METRO model uses a single-scale graph update (SGU) unit \parencite{METRO}; other graph-based models use a mix-hop model for convolution \parencite{MCFGNN, Sriramulu, MSGNet, WuZonghan}.  Dynamic GNNs are typically applied to systems with inherently correlated features such as traffic sensor measurements \parencite{DGCRN, Liang, Ye}. While many authors exploit learned pairwise spatial interactions to improve their forecast \parencite{Huang2019, Li, Liang}, some are limited to static graphs \parencite{Kipf, Shang}. This is problematic because spatial relationships are regime-dependent and time-evolving \parencite{Saggioro, ChenWu}.

Among other methods, transformers have become a dominant paradigm in time series prediction. Variants such as DSformer, STAM, and PatchTST have extended attention mechanisms to long-range forecasting. Some works address the quadratic cost of attention \parencite{Informer, Fedformer}, and others question whether attention is an effective mechanism at all: DLinear, for example, argues that positional encodings do not adequately prevent the loss of temporal information \parencite{Dlinear}. Recent benchmarking studies suggest that while transformers possess high intrinsic capacity, their inductive bias is weak for dynamical systems and they require substantial computational resources \parencite{Gilpin2023ModelSV}.

In contrast to transformers and GNNs, LSTMs have a reasonable track record of effectiveness for long-horizon prediction in chaotic systems \parencite{Sangiorgio, Cestnik, Uribarri, Pathak}, including forecasting bifurcations \parencite{Zhuge}. Although simpler RNN variants exhibit stronger inductive biases for dynamical systems modeling \parencite{Köglmayr, Panahi, Patel}, Gilpin et al. \parencite{Gilpin2023ModelSV} show that LSTMs consistently outperform them on chaotic ensemble forecasting, with the former retaining advantages only in data-limited or short-horizon regimes and degrading under noise. Universal concerns with RNN-based approaches for long-range forecasting include the increased possibility of gradient explosion and the compounding of errors \parencite{Gilpin2023ModelSV}. Seq2Seq architectures mitigate the latter challenge by grounding forecasts in the input history rather than only prior predictions. RNNs augmented with attention mechanisms can focus on the most relevant time steps \parencite{DyAt}. The Temporal Fusion Transformer (TFT) builds on these advances, combining recurrent encoders, multi-head attention, and variable selection to deliver multi-horizon forecasts. The TFT achieves feature fusion by learning a weighted sum of variable embeddings \parencite{Song}. Other non-graph-based feature fusion mechanisms in the literature include spatial attention \parencite{CaoMAFS,MDA,DA-RNN} and more advanced channel mixing mechanisms such as frequency-domain mixing \parencite{SiMBA} and correlation-aware selective fusion \parencite{CMMamba}.

\section{Methodology}

\begin{figure}
    \centering
    \fbox{\includegraphics[width=1.0\linewidth]{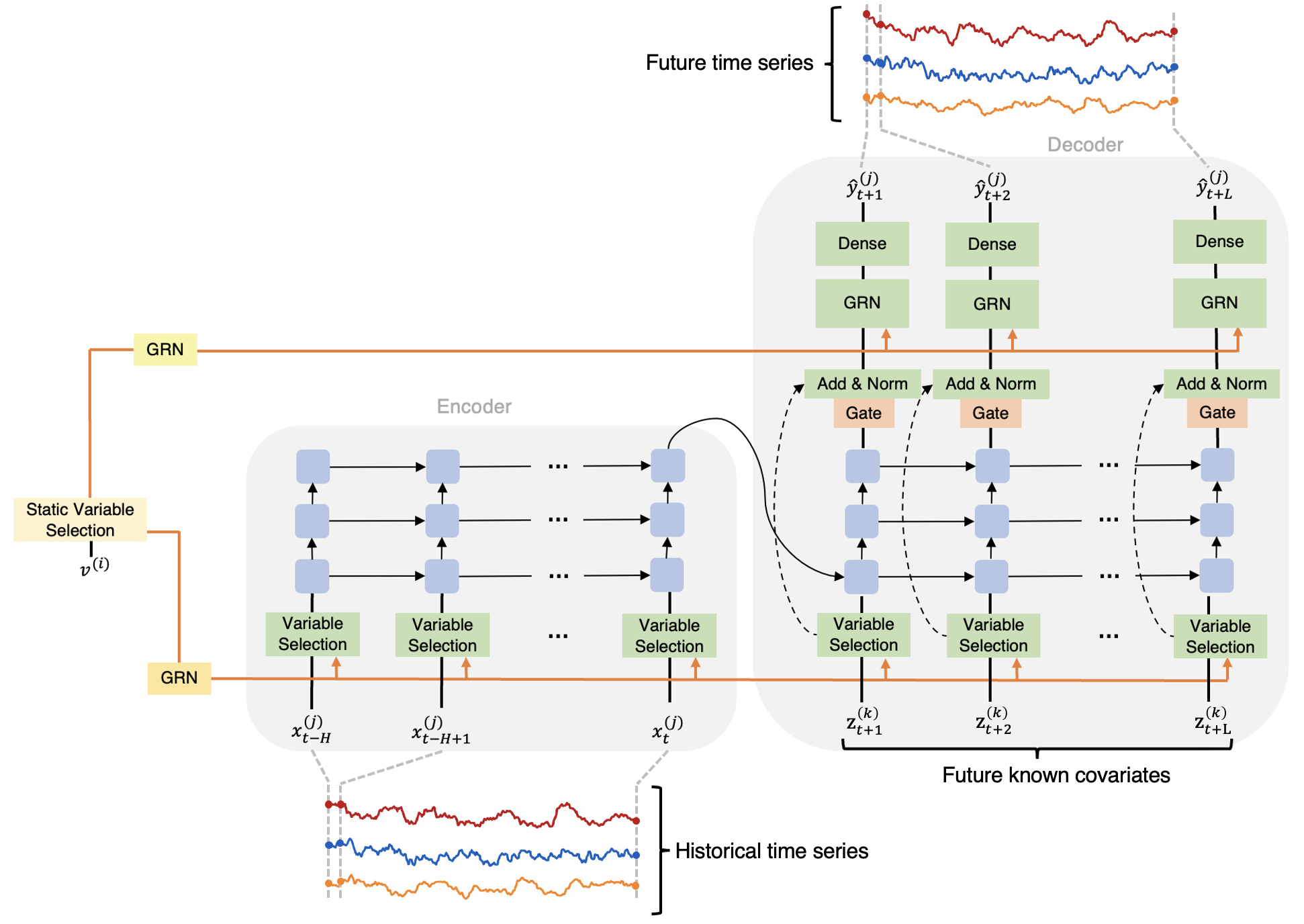}}
    \caption{Modified TFT architecture, without self-attention. $x_t^{(j)}$ denotes the $j$\textsuperscript{th} variable at time $t$, $v^{(i)}$ the $i$\textsuperscript{th} static variable, $z^{(k)}$ the $k$\textsuperscript{th} future known covariate, and $\hat{y}_t^{(j)}$ the forecasted value of the $j$\textsuperscript{th} variable at time $t$.}
    \label{architecture}
\end{figure}

We formulated tipping point prediction as a multivariate sequence forecasting problem. Given 1.) an input window of $H$ time steps of simulated state variables, external forcings, and additive noise realizations, and 2.) a set of static covariates describing model parameters and initial conditions, we trained a dynamics-informed Temporal Fusion Transformer (TFT) \parencite{TFT} to predict the future multivariate trajectory over a horizon of $L$ time steps. Long-term trajectories were then forecasted via autoregressive (sequence to sequence) rollout, with a collapse defined as the earliest crossing of the overturning strength below zero. The architecture is visualized in Figure \ref{architecture}. For the four-box surrogate, only Atlantic variables were predicted.

Training data were generated by numerically integrating the four- and six-box model equations with a fixed time step of 0.25 years. To induce stochastic variability in transition timing, additive Gaussian perturbations were introduced to the baseline values of each freshwater flux term (South, North Atlantic, North Pacific, and inter-basin) at each time step:

$$F_w^{i}(t) = F_w^{i,\text{base}} + \xi(t), \quad \xi(t) \sim \mathcal{N}(0, \sigma^2)$$

where $F_w^{i,\text{base}}$ denotes the baseline freshwater flux for basin $i$, and $\xi(t)$ represents white noise with standard deviation $\sigma$. Perturbations were sampled independently across time and across ensemble realizations, with mean zero and standard deviation $\sigma = 10^5 \: \rm{m}^3/\rm{s}$. The noise amplitude was selected such that stochastic fluctuations were small relative to the mean flux but sufficiently large to induce variability in collapse timing over multi-century horizons. This stochastic forcing represents unresolved variability and acts as an exogenous driver rather than a state-dependent perturbation.

Parameters and initial conditions were sampled uniformly within physically plausible bounds. For the six-box model, this comprised 22 static covariates including Ekman fluxes, initial pycnocline depths, and baseline freshwater fluxes for each basin. For each parameter combination, we ran the model for 1,000 years, yielding time series of 4,000 seasonal time steps. 

The four-box dataset consists of 6,000 training simulations, 401 test simulations, and 401 validation simulations. The splits contain collapse and non-collapse examples in equal proportions. For the six-box model, 1.4 million simulations were generated. To focus the surrogate on tipping dynamics, 24,909 simulations that exhibited at least one collapse within 1000 years were selected as the training simulations. The corresponding validation and test sets consist of 655 and 656 out-of-sample trajectories, respectively.

The TFT ingests three input channels: static covariates, historical time-varying covariates (Atlantic/Pacific overturning, volume fluxes, pycnocline depths, inter-basin transports, temperature, salinity), and future known time-varying covariates. When emulating deterministic simulations, no future known time-varying covariates were available, and two-dimensional sinusoidal time encodings were used to drive the decoder. In the stochastic case, the additive freshwater perturbations were treated as future known covariates. During training, the decoder received the true noise sequences used by the simulator over the prediction horizon, allowing the surrogate to condition its forecasts on the realized stochastic forcings. During inference, random noise realizations were sampled and fed to the decoder as future known covariates to produce ensemble forecasts. Thus, during training the surrogate primarily encountered diversity across parameter and initial condition space; during inference, we additionally probed its ability to generalize across noise realizations. 

To improve interpretability and suppress irrelevant inputs, variable selection networks compute relevance scores at each time step. For time-varying inputs, a gated residual network (GRN) produces relevance weights for each input feature. The weights are then normalized via a soft-max and used to form a weighted aggregation of variable-wise embeddings. Variable selection is applied separately to historical and future inputs, with weights conditioned on the aggregated static covariate embedding. Static covariates themselves pass through a variable selection GRN to produce a context vector used not only to condition variable selection but also to guide the initialization of encoder/decoder RNNs.

The backbone of the TFT is a multi-layer LSTM encoder-decoder. The encoder summarizes past dynamics by processing the aggregated historical input sequence and static embedding into a final hidden state.  The decoder then generates predictions over a specified horizon, driven by temporally indexed embeddings of future known covariates. In the original TFT, the encoder and decoder outputs are passed into a self-attention block. Each output is directly passed through a gated residual network (GRN) and a shared linear output layer, producing the forecasted values for each target variable at each future time step.

Relative to the original TFT formulation \cite{TFT}, we introduced three main modifications:

\begin{enumerate}
 \item{\textbf{Removal of the self-attention layer}. For the long sequences considered here (up to 4,000 time steps), we found that the self-attention layer increased computational cost without providing consistent performance gains. We therefore omitted the self-attention block and used the LSTM encoder-decoder plus gated residual networks and variable selection as the core architecture.}
 \item{\textbf{Alternative loss functions}. For the four-box surrogate, the native quantile loss function of the TFT was used, with only the median predicted. When applied to the six-box dataset, the same loss function produced surrogates that frequently distorted abrupt shifts and exhibited unrealistic confidence in collapse timing. Improvements were observed when switching to a soft-DTW loss function \parencite{softDTW} with the gamma parameter set to $1.0$. Soft-DTW is a differentiable relaxation of dynamic time warping. Compared to MSE, which heavily penalizes even minor phase misalignments, soft-DTW provides a smoother loss surface by softly aligning predicted and target sequences. This is particularly advantageous when forecasting MOC-related variables, where the precise timing of transitions may vary slightly across samples but the overall structure remains consistent.
 \item{\textbf{Treatment of stochastic forcing.} In the stochastic experiments, the additive freshwater flux perturbations were fed to the model as additional time-varying covariates, allowing the surrogate to condition its forecasts on hypothetical noise sequences.}
}
\end{enumerate}

Architecturally, the four-box surrogate used two LSTM layers with state size 64, while the six-box surrogate used 3 layers with state size 128. The four-box surrogate employed input and output horizons of 100 and 50 steps (25 and 12.5 years), respectively, and was trained for 19 epochs with early stopping (patience 5) with a batch size of 128. The six-box surrogate used input and output windows of 200 and 100 steps, and was trained for 10 epochs (patience 3) and batch size 32. Both models used a dropout rate of 20\%. All variables were standardized to zero mean and unit variance using statistics computed on the training set. We used the Adam optimizer with a learning rate of $1 \times 10^{-4}$, no weight decay, and gradient clipping at a norm of 1.

\subsection{Benchmark Comparison}

To contextualize the performance of the surrogate, we benchmark it against five state-of-the-art architectures:

\begin{itemize}
\item{\textbf{SegRNN} \cite{SegRNN}: SegRNN adapts recurrent architectures specifically for long lookback and forecast horizons by replacing point-wise recurrence with segment-wise iterations and parallel multi-step forecasting. The idea is to reduce recurrent iterations and improve efficiency compared to conventional RNNs.}
\item{\textbf{iTransformer} \cite{iTransformer}: A variant of the Transformer tailored for multivariate time series that inverts the conventional tokenization: instead of treating time steps as tokens, each variable's history is embedded as a token. This structure enhances learning of cross-variable relationships and long-range dependencies while leveraging self-attention for multivariate correlation.}
\item{\textbf{TiDE} \cite{TiDE}: A forecasting model that combines MLP encoders and decoders to summarize past data and generate long-horizon predictions efficiently. TiDE aims to retain the simplicity of linear models while modeling non-linear structures and covariates, often matching or outperforming more complex models on standard benchmarks with lower training cost.}
\item{\textbf{Mamba} \cite{Mamba}: A state-space model that integrates structured state space modules to capture long-range dependencies with linear computational complexity.}
\item{\textbf{TimeXer} \cite{TimeXer}: A transformer-derived forecasting model that enhances attention mechanisms for capturing both temporal dependencies and multivariate correlations across time series segments. TimeXer builds on patching and variate-oriented attention strategies to improve long-term multivariate forecasting performance compared to standard transformer baselines.}
\end{itemize}

All models were trained with identical data preprocessing, input/output window sizes, and hyperparameters adapted from Jena Climate dataset benchmarks (96-step windows) \parencite{jena_climate_dataset}. Static covariates were provided as constant time series (only the TFT has a separate mechanism for integrating static covariates). We evaluated under two experimental conditions:

\begin{enumerate}
\item{With future known freshwater fluxes, where exogenous forcings are provided to the decoder, and}
\item{Without future known covariates, where the models must extrapolate dynamics without access to future forcing information.}
\end{enumerate}

Each model was trained using both MSE and soft-DTW losses, to isolate the effect of alignment-aware objectives. Performance was assessed using single-prediction error (RMSE (1)), auto-regressive rollout error (RMSE (AR)), soft-DTW alignment error, correlation of predicted and true Atlantic collapse timing, correlation of end-state Atlantic overturning strength, and the percentage of collapse events correctly predicted. Results are presented in Tables \ref{BenchmarkResultsWithFutureFluxes} and \ref{BenchmarkResultsWithoutFutureFluxes}.

\section{Results}

\begin{figure}
    \centering
    \fbox{\includegraphics[width=1.0\linewidth]{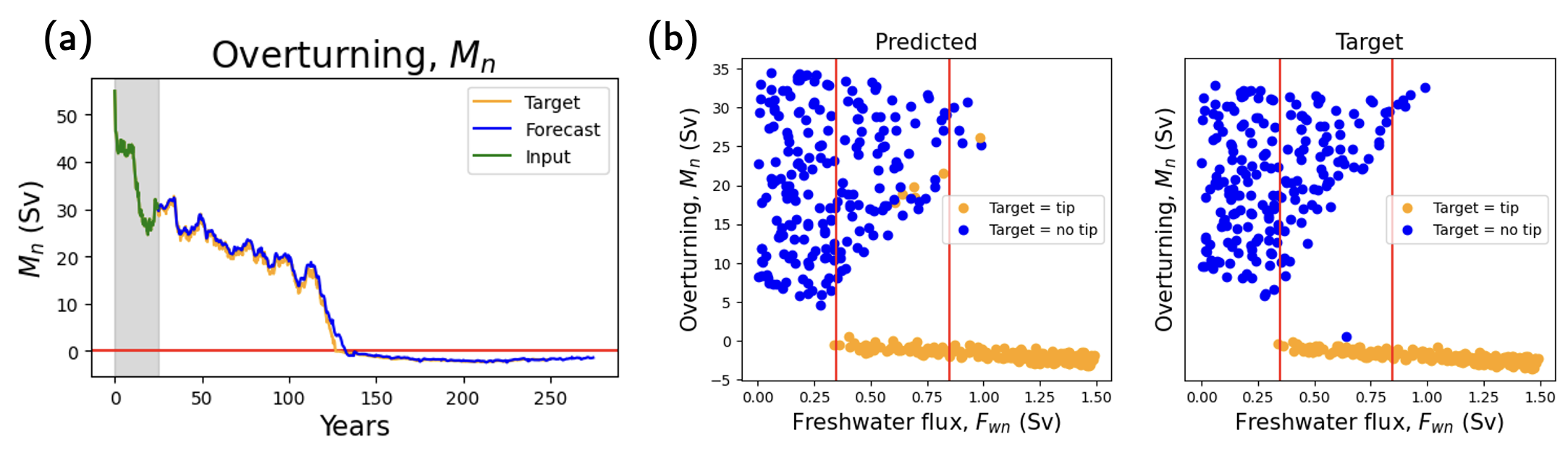}}
    \caption{(a) Atlantic overturning forecast result for a representative sample from the stochastic four-box dataset. Green is the 100 time step input observation, with each time step representing a quarter year; blue is the auto-regressive sequence-to-sequence forecast given 50 time-step prediction horizons; yellow is the ground truth. (b) Atlantic overturning, $M_n$ versus freshwater flux, $F_w^n$, for 401 test samples. Yellow indicates that the stochastic four box model forecasted a tip; blue indicates no tip. For a small subset of samples, the model predicts a false-positive tip. The vertical red lines demarcate the stable state boundaries in the underlying bifurcation diagram for non-stochastic simulations.}
    \label{FourBoxResult}
\end{figure}

\begin{figure}
    \centering
    \fbox{\includegraphics[width=1.0\linewidth]{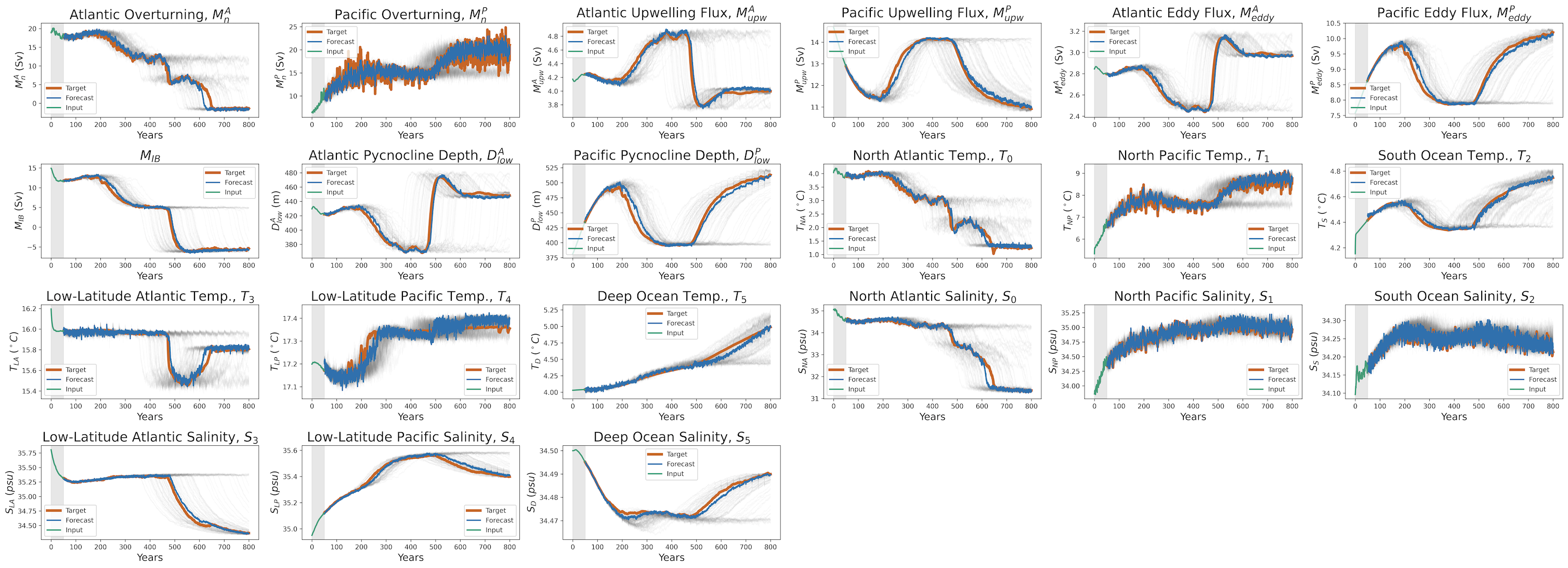}}
    \caption{Autoregressive rollouts for all predicted variables on a representative test trajectory (median RMSE across the test set). Model predictions (blue) are compared against the ground-truth trajectory (orange) generated from a fixed realization of stochastic forcing. Grey trajectories show 100 rollouts under randomized noise realizations, illustrating uncertainty in transition timing.}
    \label{MedianTestExperiment}
\end{figure}

\begin{figure}
    \centering
    \fbox{\includegraphics[width=1.0\linewidth]{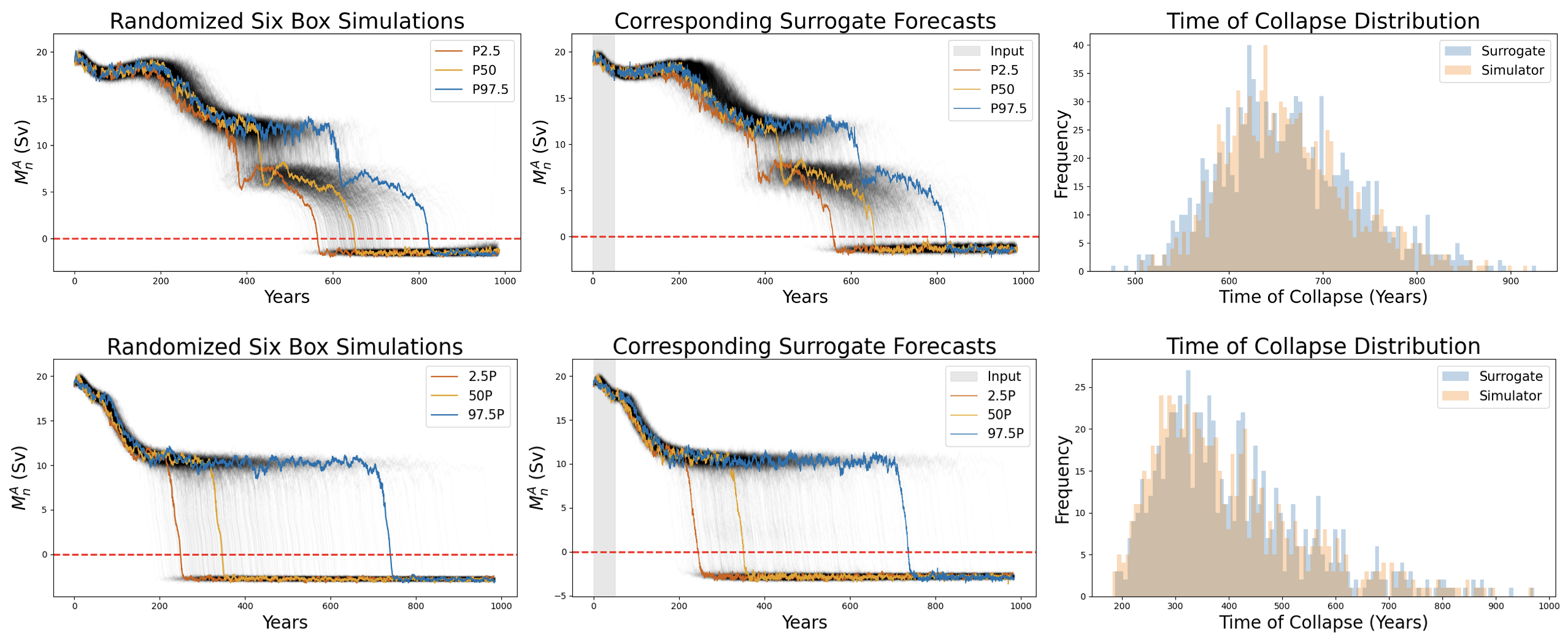}}
    \caption{Two examples (top, bottom) demonstrating ensemble fidelity of surrogate forecasts for stochastic Atlantic overturning collapse in the six-box model. Left: 1,000 stochastic realizations of the six-box simulator initialized from identical initial conditions and parameters but driven by independent freshwater flux perturbations. Thin gray lines denote individual trajectories of Atlantic overturning strength, $M_n^A$; colored curves indicate the 2.5th (P2.5), 50th (P50), and 97.5th (P97.5) percentiles in collapse time across realizations. The red dashed line marks the collapse threshold at $M_n^A = 0 \text{Sv}$. Center: Corresponding surrogate forecasts initialized using the first 50 years of each trajectory (shaded region). Forecasts were rolled forward auto-regressively for the remaining 800 years under the randomized perturbations of the corresponding simulations. Percentile bands were computed analogously to the simulator. Right: Histogram of collapse times across ensemble members for the simulator (orange) and surrogate (blue). The surrogate's timing statistics demonstrate distributional alignment in transition timing.}
    \label{SixBoxAtlanticCollapseResult}
\end{figure}

\begin{figure}
    \centering
    \fbox{\includegraphics[width=0.75\linewidth]{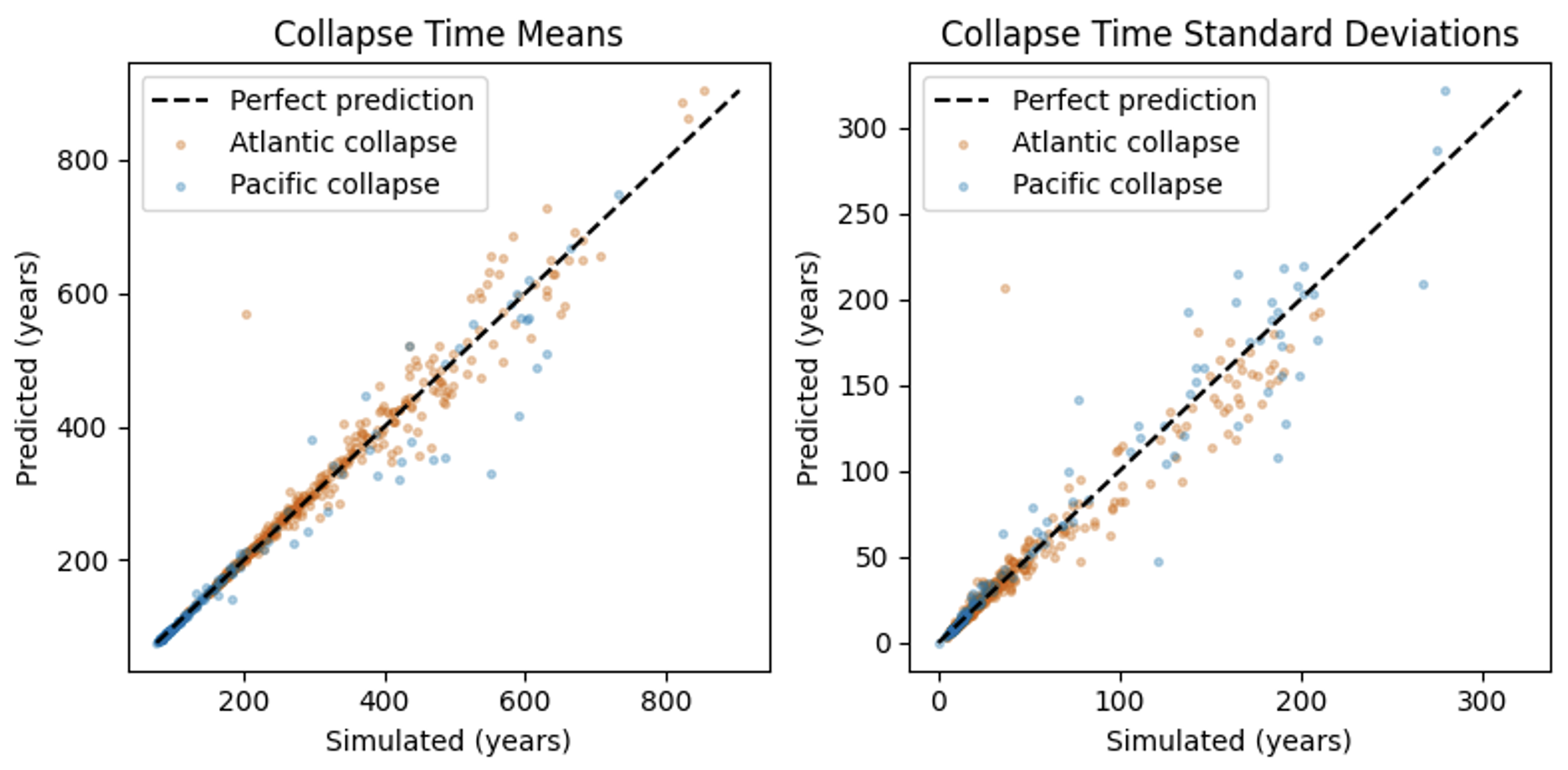}}
    \caption{Parity plots of predicted versus true transition timing for all 389 Atlantic collapse examples and 175 Pacific collapse examples from the six-box test dataset.}
    \label{SixBoxParityPlots}
\end{figure}

\subsection{Four-Box Model}

Figure \ref{FourBoxResult}(a) shows a representative auto-regressive forecast of Atlantic overturning circulation ($M_n$) for a single test trajectory. The model observes 100 time steps (25 years) of historical data and generates 20 sequential forecasts of 50 time steps each (12.5 years per forecast), rolled forward by recursively feeding predictions back as inputs. A collapse event occurs near year 140 as $M_n$ dips below $0$ Sv, indicating a transition from deep circulation to shallow (reversed) circulation. The surrogate correctly anticipated this transition with a timing error of about 10 years (equivalent to the $8^{\text{th}}$ forecast window out of 20 total).

Figure \ref{FourBoxResult}(b) shows the relationship between the final predicted (left) and true (right) overturning strength in year 1000 versus the baseline freshwater flux parameter $F_w^n$ for all 401 test samples. The points are colored by whether a collapse occurred (yellow) or not (blue). The surrogate correctly predicted collapse in 395/401 cases (98.5\%). In the six false-negative cases, the model's trajectory remained in the high-overturning regime despite the ground truth exhibiting collapse. 

\subsection{Six-Box Model}

Across all 389 Atlantic collapse examples in the six-box test dataset, the surrogate correctly identifies collapse in 97.7\% of cases. Among these, predicted and true collapse times co-vary with a Pearson correlation coefficient of 0.934.

Figure \ref{MedianTestExperiment} shows a representative autoregressive rollout for all predicted variables on a test trajectory with median long-term RMSE. The model closely tracks the target evolution across variables, maintaining stability over multi-century horizons. Ensemble rollouts reveal sensitivity to perturbations in the vicinity of collapse. To assess whether the surrogate reproduces the distribution of transition times under stochastic forcing, an ensemble experiment was conducted. For each parameter and initial condition in the test set that led to collapse, 1,000 simulations were generated using identical initial conditions but independent noise sequences over 4,000 time steps. Corresponding surrogate predictions were initialized from the same 50-year observation window and rolled forward autoregressively for 950 years. Figure \ref{SixBoxAtlanticCollapseResult} (left) shows simulator trajectories for a representative test case. While individual realizations (gray) vary in collapse timing, they converge to a common collapsed state. The colored curves represent the 2.5th, 50th (median), and 97.5th percentiles of collapse time for the simulated data, approximating a two-standard-deviation interval. Figure \ref{SixBoxAtlanticCollapseResult} (center) shows corresponding surrogate forecasts. Figure \ref{SixBoxAtlanticCollapseResult} (right) compares the resulting distributions of collapse times, demonstrating close agreement between the surrogate (blue) and simulator (orange).

Figure \ref{SixBoxParityPlots} extends this analysis to all test examples exhibiting collapse. For each parameter and initial condition combination, the mean and standard deviation of collapse time across 1,000 noise realizations are computed. The left panel shows the predicted versus true mean collapse time ($r=0.987$ for Atlantic; $0.977$ for Pacific); the right panel shows predicted versus true standard deviation of collapse time ($r=0.964$ for Atlantic; $0.971$ for Pacific). The surrogate closely captures both first- and second-order statistics of the transition time distributions.

\subsection{Model Comparison}

\begin{table}[!ht]
    \centering
    \caption{Benchmark performance with future known freshwater fluxes provided as covariates to the decoder. Models are evaluated on single-step prediction error, auto-regressive rollout error, alignment error (soft-DTW), and their ability to capture Atlantic collapse timing and end-state overturning strength. Access to future forcing information enables conditioning on exogenous drivers during forecasting.}
    \label{BenchmarkResultsWithFutureFluxes}
    \resizebox{\textwidth}{!}{
    \begin{tabular}{|l|l|l|l|l|l|l|l|}
    \hline
    \textbf{Model} & \textbf{Training Loss} & $\mathbf{SDTW\ (1)}$ & $\mathbf{RMSE\ (1)}$ & $\mathbf{RMSE\ (AR)}$ & $\mathbf{RMSE_{M_n^A}\ (AR)}$ & $\mathbf{r_{collapse,A}}$ & $\mathbf{r_{M_n^A, end}}$ \\ \hline
    TFT (original) & SDTW & 3.638 & 0.04115 & 0.1721 & 0.2241 & 0.9209 & 0.9711 \\ 
        ~ & MSE & 4.291 & \textbf{0.03876} & 0.1635 & 0.2174 & 0.8715 & 0.9659 \\
        TFT (ours) & SDTW & \textbf{2.831} & 0.0408 & \textbf{0.1604} & \textbf{0.2015} & \textbf{0.934} & \textbf{0.9763} \\ 
        ~ & MSE & 3.691 & 0.03992 & 0.1693 & 0.2275 & 0.8874 & 0.9716 \\ 
        TiDE & SDTW & 32.73 & 0.125 & 0.6828 & 1.148 & N/A & 0.6733 \\ 
        ~ & MSE & 33.46 & 0.1065 & 0.6718 & 1.141 & N/A & 0.6787 \\ 
        Persistence & N/A & 44.09 & 0.1206 & 0.734 & 1.256 & N/A & 0.6654 \\ \hline
    \end{tabular}}
\end{table}

\begin{table}[!ht]
    \centering
    \caption{Benchmark performance without access to future known covariates. Models rely solely on historical inputs and positional encodings to extrapolate system dynamics. Metrics are identical to Table \ref{BenchmarkResultsWithFutureFluxes}.}
    \label{BenchmarkResultsWithoutFutureFluxes}
    \resizebox{\textwidth}{!}{
    \begin{tabular}{|l|l|l|l|l|l|l|l|}
    \hline
    \textbf{Model} & \textbf{Training Loss} & $\mathbf{SDTW\ (1)}$ & $\mathbf{RMSE\ (1)}$ & $\mathbf{RMSE\ (AR)}$ & $\mathbf{RMSE_{M_n^A}\ (AR)}$ & $\mathbf{r_{collapse,A}}$ & $\mathbf{r_{M_n^A, end}}$ \\ \hline
        TFT (original) & SDTW & 6.879 & 0.05392 & 0.2428 & 0.3379 & 0.7903 & \textbf{0.9697} \\ 
        ~ & MSE & 7.834 & 0.05311 & 0.2185 & 0.2929 & 0.7736 & 0.9512 \\
        TFT (ours) & SDTW & \textbf{6.745} & 0.05456 & 0.2018 & 0.2715 & \textbf{0.8600} & 0.9659 \\
        ~ & MSE & 7.952 & \textbf{0.05248} & \textbf{0.1948} & \textbf{0.2659} & 0.8451 & 0.962 \\
        TiDE & SDTW & 32.74 & 0.125 & 0.7072 & 1.188 & N/A & 0.664 \\
        ~ & MSE & 33.46 & 0.1065 & 0.6959 & 1.1793 & N/A & 0.6689 \\
        Mamba & SDTW & 12.73 & 0.07552 & NaN & NaN & 0.1729 & NaN \\
        ~ & MSE & 13.12 & 0.0681 & NaN & NaN & 0.294 & NaN \\
        iTransformer & SDTW & 7.3071 & 0.08046 & 0.9051 & 1.553 & 0.389 & -0.2978 \\
        ~ & MSE & 9.54 & 0.05777 & 0.5924 & 0.8224 & 0.2303 & 0.635 \\
        SegRNN & SDTW & 13.26 & 0.1025 & 3.543 & 0.8624 & 0.1031 & 0.6881 \\
        ~ & MSE & 15.43 & 0.0726 & 1.269 & 1.708 & 0.2516 & 0.4071 \\
        TimeXer & SDTW & 11.63 & 0.07415 & 1568000 & 2.666 & 0.377 & 0.2144 \\
        ~ & MSE & 14.08 & 0.0698 & 21849 & 1.91 & 0.09379 & -0.01336 \\
        Persistence & N/A & 44.09 & 0.1206 & 0.734 & 1.256 & N/A & 0.6654 \\ \hline
    \end{tabular}}
\end{table}

Across both setups, the modified TFT variant consistently achieves the lowest alignment scores and competitive RMSE metrics, particularly in auto-regressive rollouts. Notably, the soft-DTW objective improves the transition timing fidelity compared to MSE alone, particularly when exogenous forcing is available. In contrast, several of the alternatives exhibit strong short-horizon performance but degrade under rollout, with most struggling to maintain stability and even predict a collapse. Divergent trajectories are reflected in invalid metrics (NaN) or extremely high RMSE, highlighting the difficulty of maintaining stability over multi-century auto-regressive rollouts, rather than necessarily reflecting inherent flaws in the architectures themselves.

\subsection{Computational Benchmarking}

When sampling uniformly at random within the estimated bounds for 21 parameters, about 1.5\% of thousand-year simulations involved a tip (Atlantic or Pacific collapse). Under these odds, generating only 1,000 tip examples would require around 67,000 simulations. Running 67,000 simulations with TFT would take 4 minutes within the memory constraints of a single H100 GPU (4.5 minutes with an A100), whereas running an equivalent number of simulations with the 6 box numerical simulator would take 31 hours on a single CPU core (or 4 minutes across 472 CPU cores). Even 1,000 tip examples is small compared to the number of samples required to characterize the uncertainty in tipping time across multiple noise realizations and multiple combinations of parameters and initial conditions. As noted previously, the reduced 6 box model is already orders of magnitude faster than global climate models with full ocean dynamics. Studying tipping point behavior is challenging for this reason. 

\section{Discussion}

While deterministic models of the Earth system could in principle allow exact forecasts, the butterfly effect in chaotic dynamics limits the predictability horizon; after some time, small uncertainties begin to grow exponentially \parencite{Datseris2022}. Due to this effect and the stochastic nature of the simulations in this study, the surrogate model cannot emulate the ground truth to arbitrary precision. Given these limitations, the fidelity of the surrogate-emulated predictions is surprisingly high. The close agreement in both mean and standard deviation of collapse time across ensembles (Figure~\ref{SixBoxParityPlots}) suggests that the model has learned a robust mapping from parameter and initial condition space to the induced distribution over collapse times. Producing ensembles that diverge in timing while converging to the same final state reflects an implicit understanding of the geometry of the dynamics in state space.

Overall, the results add to the empirical evidence that LSTMs may be suited to chaotic systems forecasting. Cestnik and Abel \parencite{Cestnik} demonstrated that LSTMs trained on oscillatory data can generalize outside the training domain and respond coherently to interventions, hinting at a latent understanding of system dynamics. Uribarri and Mindlin \parencite{Uribarri} showed that sequence-to-sequence LSTMs trained on chaotic trajectories, given a sufficiently long prediction horizon, can reconstruct a hidden state space topologically equivalent to the true attractor. The authors draw a connection to Takens’ Theorem, which states that time-delay embeddings with sufficient lags can recover a diffeomorphic image of the original system's attractor. Because LSTMs have adaptive memory, there are no theoretical guarantees that they implicitly perform delay embeddings of the input signal; however, they can exhibit such fading-memory behavior in practice, potentially explaining their success on chaotic prediction tasks.

The computational advantages of the surrogate have implications for parameter inference and uncertainty quantification. Parameter and initial condition inference typically requires numerous forward model evaluations---MCMC, SMC, iterated EnKF, and variational data assimilation methods often demand thousands to millions of model runs. The box model becomes prohibitively expensive for these ensemble-based workflows and is not natively differentiable. The $465\times$ speedup achieved by the surrogate expands the scope of tractable analyses. Beyond speed, differentiability with respect to parameters and initial conditions enables gradient-based data assimilation and sensitivity analysis algorithms that exploit gradient information to converge more efficiently than gradient-free methods.

Generating the training and evaluation datasets for the surrogate required 40 hours of simulation time on 16 CPU cores. While manageable for initialization, this rate of data generation would make iterative refinement via ensemble-based data assimilation computationally impractical if each iteration required retraining. However, because the surrogate is conditioned on model parameters and initial conditions as inputs, new simulations prescribed by data assimilation workflows can be incorporated to enrich the training dataset and refine the surrogate incrementally, rather than requiring retraining from scratch. This is particularly valuable for active learning strategies that adaptively sample parameter space to improve surrogate fidelity. The combination of computational efficiency, differentiability, and conditioning on system parameters positions learned surrogates as enabling tools for inverse problems when simulators remain computationally prohibitive. 

\section{Acknowledgments}

This work was supported by NASA under Grant No. 80NSSC25K0062. Any opinions, findings, conclusions or recommendations expressed in this material are those of the author and do not necessarily reflect the views of NASA. This work was also supported by independent research and development funding from the Research and Exploratory Development Mission Area of the Johns Hopkins Applied Physics Laboratory.

% \bibliographystyle{unsrtnat}
% \bibliography{references}
\printbibliography
\end{document}